\documentclass{article} 
\usepackage{iclr2025_conference,times}

\usepackage{graphicx}
\usepackage[utf8]{inputenc} 
\usepackage[T1]{fontenc}    
\usepackage{hyperref}       
\usepackage{url}            
\usepackage{booktabs}       
\usepackage{amsfonts}       
\usepackage{nicefrac}       
\usepackage{microtype}      
\usepackage{xcolor}         
\usepackage{svg}
\usepackage{amsmath}
\usepackage{multirow}
\usepackage{array}
\usepackage{geometry}
\usepackage{colortbl}
\usepackage{caption}
\usepackage{quoting}
\usepackage{subcaption}
\usepackage{listings}       
\quotingsetup{leftmargin=1cm}
\usepackage{enumitem}
\usepackage[draft]{changes} 
\usepackage{float}

\definechangesauthor[name={Parth Shah}, color=blue]{ps}

\newcommand{\fig}[1]{Fig.~\ref{#1}}

\makeatletter
\renewcommand\@makefnmark{\textsuperscript{\dagger}}
\renewcommand\@fnsymbol[1]{\dagger}
\makeatother

\title{How Many Instructions Can LLMs Follow at Once?}

\author{
Daniel Jaroslawicz$^{1}$ \quad Brendan Whiting$^{1}$ \quad Parth Shah$^{1}$ \quad Karime Maamari$^{1}$\\
$^1$Distyl AI\\
\texttt{\{daniel, brendan, parth, karime\}@distyl.ai}
}

\begin{document}
\maketitle

\begin{abstract}
Production-grade LLM systems require robust adherence to dozens or even hundreds of instructions simultaneously. However, the instruction-following capabilities of LLMs at high instruction densities have not yet been characterized, as existing benchmarks only evaluate models on tasks with a single or few instructions. We introduce IFScale, a simple benchmark of 500 keyword-inclusion instructions for a business report writing task to measure how instruction-following performance degrades as instruction density increases. We evaluate 20 state-of-the-art models across seven major providers and find that even the best frontier models only achieve 68\% accuracy at the max density of 500 instructions. Our analysis reveals model size and reasoning capability to correlate with 3 distinct performance degradation patterns, bias towards earlier instructions, and distinct categories of instruction-following errors. Our insights can help inform design of instruction-dense prompts in real-world applications and highlight important performance-latency tradeoffs. We open-source the benchmark and all results for further analysis at \url{https://distylai.github.io/IFScale}.
\end{abstract}



\begin{figure}[!htbp]
\centering
\setlength{\abovecaptionskip}{4pt}  
  \setlength{\belowcaptionskip}{-4pt} 
\includegraphics[
width=0.8\textwidth
]{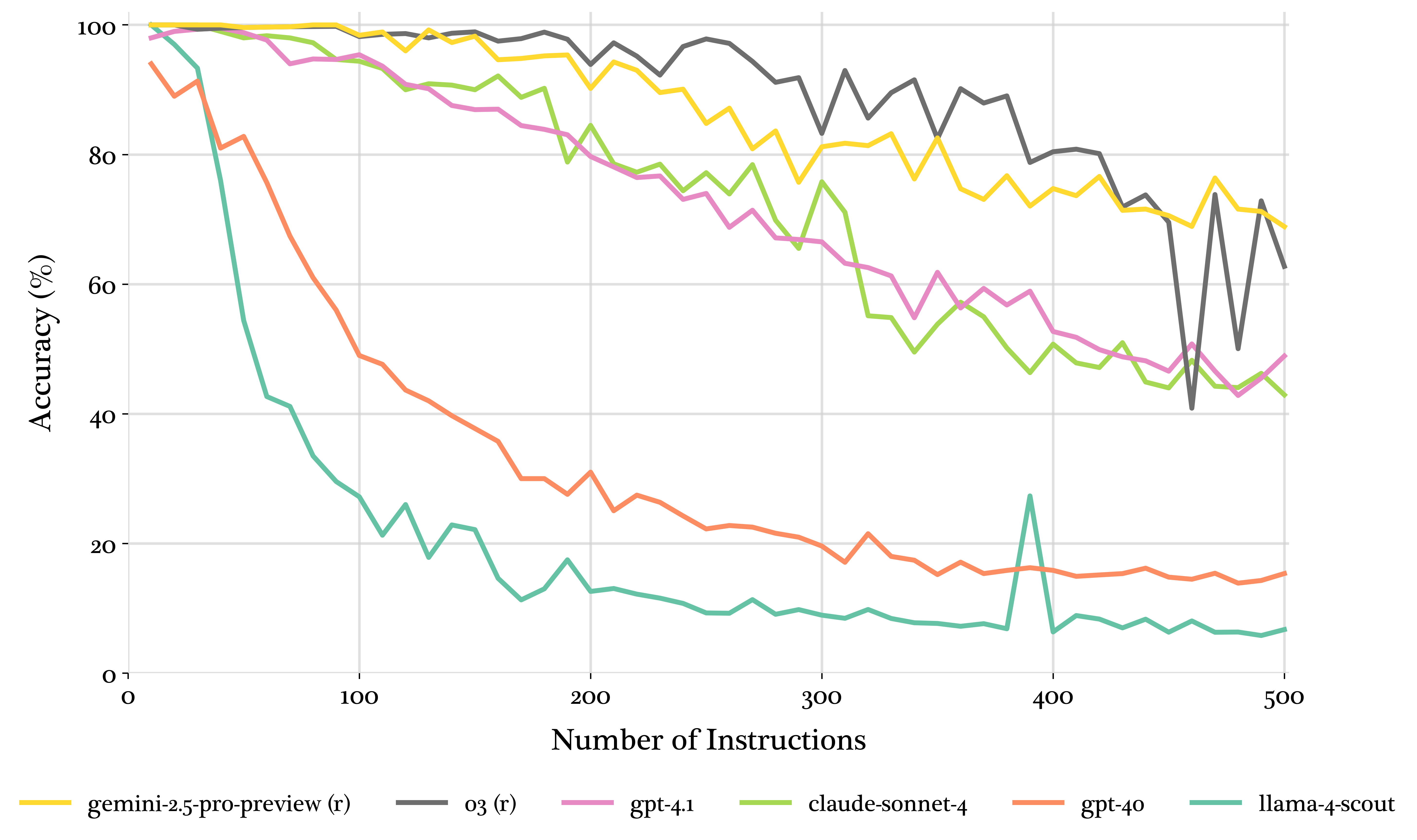}
\caption{Model instruction-following accuracy across increasing densities, averaged over 5 runs. Three distinct degradation patterns emerge: (1) threshold decay---near-perfect performance until a critical density, then rising variance and decreased adherence (reasoning models like \texttt{o3}, \texttt{gemini-2.5-pro}), (2) linear decay (\texttt{gpt-4.1}, \texttt{claude-sonnet-4}), and (3) exponential decay (\texttt{gpt-4o}, \texttt{llama-4-scout}).}
\label{fig:accuracy_all}
\end{figure}

\section{Introduction}

As large language models (LLMs) are increasingly being deployed in production systems requiring precise specification adherence, understanding their limitations is essential for reliable operation \citep{ouyang2022traininglanguagemodelsfollow, sanh2022multitaskpromptedtrainingenables, wei2022finetunedlanguagemodelszeroshot, song2025dynamicsinstructionfinetuningchinese}. From content generation systems that must adhere to style guidelines and factual requirements, to automated workflows that integrate dozens of business rules and compliance standards, to agentic systems requiring robust memory layers and tool usage, modern applications demand models that can execute complex tasks while satisfying multiple simultaneous instructions \citep{delangis2024dynamicmultirewardweightingmultistyle, kulkarni2025agentsllmagenticworkflow, xu2025amemagenticmemoryllm}. Real-world failures, such as chatbots inventing non-existent policies or providing misleading advice, highlight the operational and legal risks of imperfect instruction following.

This challenge has become pressing as recent advances dramatically expand what we can feasibly ask models to handle. Context windows have grown from thousands to millions of tokens \citep{geminiteam2024gemini15unlockingmultimodal}, and reasoning capabilities over extended contexts have improved \citep{openai2024openaio1card, deepseekai2025deepseekr1incentivizingreasoningcapability}. These developments theoretically enable single-call requests with many simultaneous instructions, rather than the standard paradigm requiring careful decomposition or retrieval \citep{chung2025longcontextneedleveraging, Chan_2025, maamari2024deathschemalinkingtexttosql}. To confidently move towards increased instruction density, we must first answer: how many instructions can models actually handle before performance meaningfully degrades?

Existing instruction-following benchmarks provide limited insight into this question. Early evaluations typically assessed models using simpler tasks or small numbers of instructions per request \citep{wang2022supernaturalinstructionsgeneralizationdeclarativeinstructions, mishra2022crosstaskgeneralizationnaturallanguage}. Recent benchmarks have advanced in complexity and realism \citep{maamari2024deathschemalinkingtexttosql, jiang2024followbenchmultilevelfinegrainedconstraints, madaan2023selfrefineiterativerefinementselffeedback, he2024multiifbenchmarkingllmsmultiturn, qin-etal-2024-infobench, zeng2024evaluatinglargelanguagemodels, jing2023followevalmultidimensionalbenchmarkassessing}, but still focus on scenarios with few instructions. This leaves a gap in understanding around performance degradation under the high instruction densities that expanded model capabilities now theoretically support. To address this gap, we introduce IFScale, a benchmark designed to characterize how models handle increases in cognitive load.

The main contributions of this work include: (1) \textbf{IFScale}: a benchmark for evaluating instruction-following performance as instruction density increases; (2) \textbf{Comprehensive evaluation}: an evaluation revealing performance hierarchies and degradation patterns across state-of-the-art models and a detailed exploration of instruction ordering effects, error types, and task performance under high cognitive load for all models considered.

\section{Related Work}

\subsection{Evaluation of LLM Instruction Following}
Evaluation of LLM instruction following capabilities is essential to ensure that model outputs align closely with human intentions, an area increasingly explored through recent benchmarks \citep{lou2024largelanguagemodelinstruction, zeng2024evaluatinglargelanguagemodels, liu-etal-2025-reife}. Early approaches like FLAN, InstructGPT, and large-scale benchmarks such as Super-NaturalInstructions have advanced our understanding by showcasing enhanced alignment and generalization from task-tuning \citep{wei2022finetunedlanguagemodelszeroshot, ouyang2022traininglanguagemodelsfollow, wang2022supernaturalinstructionsgeneralizationdeclarativeinstructions}. However, these benchmarks have predominantly assessed performance using simpler or smaller-scale tasks, limiting understanding of model behavior under higher-density instruction scenarios. 

\subsection{Recent Benchmarks on Complex Instruction Following}
Several recent benchmarks have advanced the complexity and realism of LLM evaluation by explicitly exploring scenarios involving multiple tasks or instructions. For instruction following, ComplexBench and FollowBench evaluate LLM performance on complex instructions composed of multiple constraints \citep{wen2024benchmarkingcomplexinstructionfollowingmultiple, jiang2024followbenchmultilevelfinegrainedconstraints}. DC-Instruct introduced methods to explicitly handle interdependent or multi-intent tasks, highlighting improvements with structured approaches \citep{Xing2024DCInstructAE}. However, these benchmarks generally fail to explore how model performance degrades in many instruction scenarios. 

\subsection{Evaluations of Instruction Complexity}
Recent studies have highlighted that instruction complexity influences model performance and robustness. For example, DIM-Bench demonstrated that LLMs are vulnerable to negative or distractor requirements \citep{hwang2025llmseasilyconfusedinstructional}. Recent work has also shown that order effects matter in instruction following, with items presented earlier receiving more attention \citep{zeng2025ordermattersinvestigateposition, liu-etal-2025-reife, wen2024benchmarkingcomplexinstructionfollowingmultiple}. Yet, these evaluations typically examine instruction complexity at low densities, without exploring larger-scale combinations of instructions. 

Our benchmark addresses these limitations by evaluating performance at increased instruction densities, providing insight into performance cliffs and degradation patterns not observable in single- or few-instruction evaluations.

\section{IFScale}

We propose IFScale, a benchmark designed to investigate how model performance degrades as instruction density increases. The task is to generate a professional business report while including a set of keywords in the output. Each instruction is a constraint to include a specific keyword in the generated report. This allows us to easily scale instruction density from $10$ to $500$ instructions with a step size of $10$ and automatically grade performance by keyword inclusion.

\subsection{Term Vocabulary Construction}
We compile a high-precision vocabulary of business-relevant one-word instructions from U.S.
SEC 10-K filings. For each filing, we prompt \texttt{o4-mini} to extract the top $500$ candidate terms as a JSON list.
Extracted lists are fuzz-match deduplicated.
A look-back validation step retains only terms found as whole words in the raw 10-K
corpus, to avoid any improper extraction by the model. 
We then filter by Zipf frequency ($\geq$  1.0) to ensure all terms exist in standard English terminology. 
Morphological variants are collapsed to a single lemma via WordNet lemmatization.
To guard against semantic redundancy, we embed all remaining terms using OpenAI's \texttt{text-embedding-3-small}, compute each term's nearest-neighbor cosine distances, and prune any term whose distance
falls below the mean. 
Finally, we estimate each lemma's generation "difficulty" by
averaging $\exp(-\operatorname{avg\_logprob})$ over three zero-temperature
\texttt{gpt-4.1-nano} completions, rank by descending perplexity, and select the top $500$ terms
as our final rule vocabulary (Appendix \ref{app:vocab_table}).

\subsection{Implementation Details}\label{sec:implementation}

For each experiment, we evaluate a grid of instruction densities $N \in \{10,20,\dots,500\}$ under five random seeds.

\begin{itemize}[leftmargin=*]
  \item \textbf{Prompt construction}: Sample $N$ keywords from the pruned vocabulary and create a list of instructions of the form: "Include the exact word \{keyword\}". Instruct the model to build a multi-section professional business report while obeying the list of instructions (Appendix \ref{app:report_prompt}). 
  \item \textbf{Retry logic}: Issue each prompt to the respective model with retries when lists of constraints ($\geq$10 comma-separated single words), response refusals ($<$20 words), or incoherent reports (validated by a secondary \texttt{o4-mini} coherence check) are identified.
  \item \textbf{Evaluation}: Evaluate each report's adherence via wildcard-enabled regex matching on the text.
\end{itemize}

\section{Experiments}

\subsection{Experimental Setup and Metrics}

We evaluated a total of 20 models spanning seven providers, as highlighted in Fig. \ref{fig:instruction_following}. We evaluated each model via the OpenRouter API using default generation parameters to maintain natural generation characteristics. We set reasoning effort to "high" where applicable. Five independent random seeds were run per instruction density level ($N\in\{10,20,\dots,500\}$), and stratified sampling ensured consistent difficulty across runs.


\subsection{Evaluation Methodology}
We assess each generated report's adherence to its instructions via deterministic pattern matching. First, we perform case-insensitive, style-insensitive exact-match searches using regular expressions to identify proper keyword inclusions. Instructions not found are counted as omission errors. Instructions matching at least an 80\%-length prefix of each term are counted as modification errors. We compute per-model and per-density omission and modification rates by aggregating across seeds. To quantify primacy effects, we partition each instruction list into early, middle, and late thirds and measure error rates within each bucket. Standard deviation is computed by taking the sample standard deviation of accuracy scores across the five random seeds at each density level.

\subsection{Performance Analysis}

\begin{figure}[t!]
    \centering
    \includegraphics[width=1.0\textwidth]{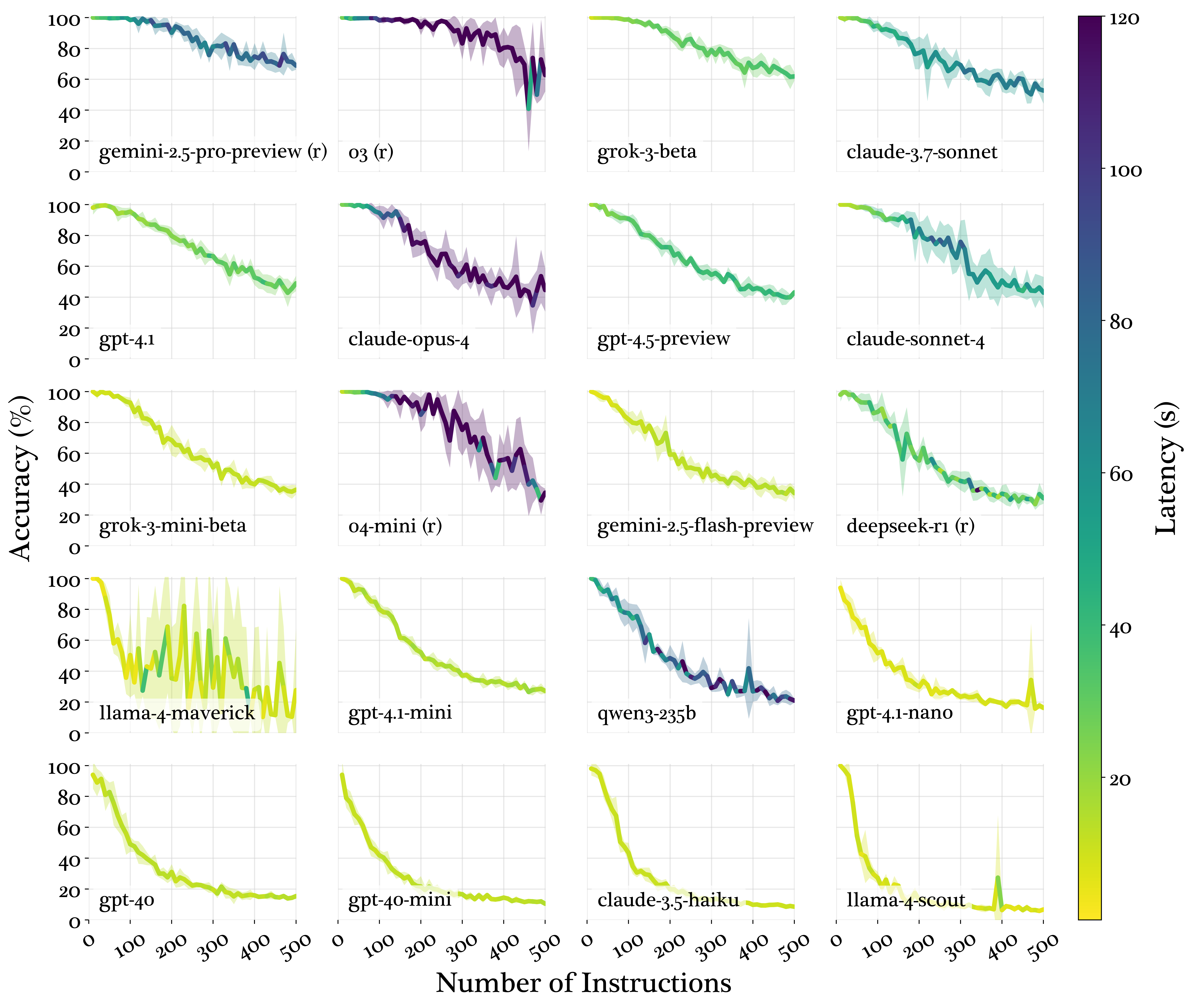}
    \caption{Model performance degradation as instruction density increases from 10 to 500 instructions, with line coloring representing generation latency and shading indicating $\pm 1$ standard deviation across five runs. Models are ordered by accuracy at 500 instructions. Larger or newer models and models with reasoning tend to outperform smaller or earlier generation models that show rapid early degradation.}
    \label{fig:instruction_following}
\end{figure}

Figure~\ref{fig:instruction_following} displays model performance patterns across the instruction density spectrum, while Appendix~\ref{app:full_results} presents comprehensive metrics across multiple dimensions.


Generally, reasoning models (indicated in all figures by "(r)" appended to model name) outperform their general-purpose counterparts. They maintain near-perfect performance through moderate densities (100-250 instructions) before degradation, suggesting that reasoning capabilities provide advantages for instruction tracking and satisfaction (see appendix \ref{app:reasoning_analysis} for further reasoning model analysis). Also, naturally, newer generation general-purpose models generally outperform their older-generation counterparts, and larger models outperform smaller counterparts.

However, several outliers challenge these general trends. \texttt{grok-3} (61.9\% at 500 instructions) approaches the performance of \texttt{o3} (62.8\%) with significantly less variance, despite not being run in reasoning mode. \texttt{claude-3.7-sonnet} outperforms the newer \texttt{claude-opus-4} and \texttt{claude-sonnet-4} at max density (52.7\% vs. 44.6\% and 42.9\% respectively). Additionally, \texttt{deepseek-r1} (30.9\%) underperforms given its reasoning model classification, and \texttt{qwen3} (26.9\%) falls short of expectations for a large, new-generation model. Finally, \texttt{gpt-4o} displays surprisingly weak performance, showing rapid decay more characteristic of small models like \texttt{gpt-4.1-nano}. 

\subsection{Degradation Pattern Analysis}

Analysis of accuracy degradation curves reveals three distinct patterns as shown in \fig{fig:accuracy_all}:

\textbf{Threshold decay}: Performance remains stable until a threshold, then transitions to a different (steeper) degradation slope and displays increased variance. The top two models (\texttt{gemini-2.5-pro}, \texttt{o3}) demonstrate this clearly, maintaining near-perfect performance through 150 or more instructions before declining. Notably, these are both reasoning models, indicating that deliberative processing architectures provide robust instruction tracking up to critical thresholds, beyond which systematic degradation occurs.

\textbf{Linear decay}: Characterized by steady, predictable decline in performance. \texttt{gpt-4.1} and \texttt{claude-3.7-sonnet} exemplify this pattern, with accuracy decreasing approximately linearly across the density spectrum.

\textbf{Exponential decay}: Characterized by rapid early degradation followed by performance stabilization at low accuracy floors. Models like \texttt{claude-3.5-haiku} and \texttt{llama-4-scout} exemplify this pattern, showing steep performance drops after minimal instruction densities before asymptotically approaching consistent low-performance baselines. Notably, all exponential decay patterns appear to level off around similar accuracy floors (7-15\%), suggesting lower bounds on instruction satisfaction. 

\subsection{Variance Patterns}

We examine performance variance across the five runs per instruction density level and observe three distinct behaviors as shown in Appendix \ref{app:efficiency}: Top performing models by accuracy (e.g. \texttt{gemini-2.5-pro}, \texttt{o3}, \texttt{grok-3-beta}) display steady increases in variance, indicating reduced reliability as instruction density increases. Mid-tier performing models (e.g. \texttt{gemini-2.5-flash}, \texttt{claude-sonnet-4}) show mid-range variance peaks in the 150-300 instruction range suggesting a critical capacity zone where performance is unstable before the model collapses under cognitive load and stabilizes at consistently poor performance. Finally, the worst performing models almost immediately decrease in variance suggesting that they are overwhelmed by even a few dozen instructions. We can infer that variance decreases as models collapse under cognitive load, and that the top performing models do not yet collapse at 500 instructions. We note that \texttt{llama-4-maverick} stands out as an extreme outlier with abnormally high variance, suggesting different instruction-processing mechanisms.

\subsection{Primacy Effects}

Primacy effects refer to the tendency of models to better satisfy instructions appearing earlier versus later in the instruction list \citep{guo2024serialpositioneffectslarge, zhou2024unibiasunveilingmitigatingllm, horowitz2025llmagentsdisplayhuman}. We compute primacy effects as the ratio of error rates in the final third of instructions to error rates in the first third of instructions. A ratio greater than 1.0 indicates that later instructions are more likely to be violated.

Primacy effects display an interesting pattern across all models: they start low at minimal instruction densities indicating almost no bias for earlier instructions, peak around 150-200 instructions, then level off or decrease at extreme densities. This mid-range peak suggests that models exhibit the most bias as they begin to struggle under cognitive load at moderate densities. However, at extreme densities (300+ instructions), primacy effects uniformly diminish across all models, with most ratios converging toward 1.0--1.5 (see Appendix \ref{app:full_results} and \fig{fig:primacy_effect_split}). This convergence indicates a shift from selective instruction satisfaction to more uniform failure patterns when completely overwhelmed, indicating an instruction saturation point. Therefore, while packing more important instructions towards the beginning of a prompt may help, it becomes a less effective strategy once extreme densities are reached.

\subsection{Efficiency Analysis}

Most production applications have some latency constraints even if they do not demand real-time interaction. We analyze generation latency and accuracy tradeoffs as instruction density increases. Reasoning models exhibit the most pronounced latency increases under cognitive load: \texttt{o4-mini} scales dramatically from 12.40s at 10 instructions to 436.19s at 250 instructions and \texttt{o3} increases from 26.30s to 219.58s at 250 instructions. In contrast, general-purpose models maintain stable latency profiles: \texttt{claude-3.5-haiku} ranges from 9.32s to 10.54s, \texttt{gpt-4o} remains between 9.29s and 13.20s (Appendix \ref{app:full_results}).

The accuracy-to-latency efficiency ratio reveals practical deployment insights that pure accuracy metrics obscure (Appendix \ref{app:efficiency}). All models show declining efficiency as instruction density increases, with universal convergence toward low ratios (0-2) at high densities. However, efficiency hierarchies differ markedly from accuracy hierarchies. Fast, smaller models like \texttt{grok-3-mini}, \texttt{gemini-2.5-flash}, and \texttt{gpt-4.1-nano} achieve the highest efficiency ratios, demonstrating that computational speed can compensate for moderate accuracy losses in time-sensitive applications. Larger reasoning models like \texttt{o3} and \texttt{gemini-2.5-pro} exhibit lower efficiency ratios than several smaller models, suggesting their computational costs may outweigh accuracy benefits for practical deployment. Notably, \texttt{grok-3} maintains a high efficiency ratio and strong accuracy performance.

\begin{figure}[!t]
\centering
\includegraphics[width=1.0\textwidth]{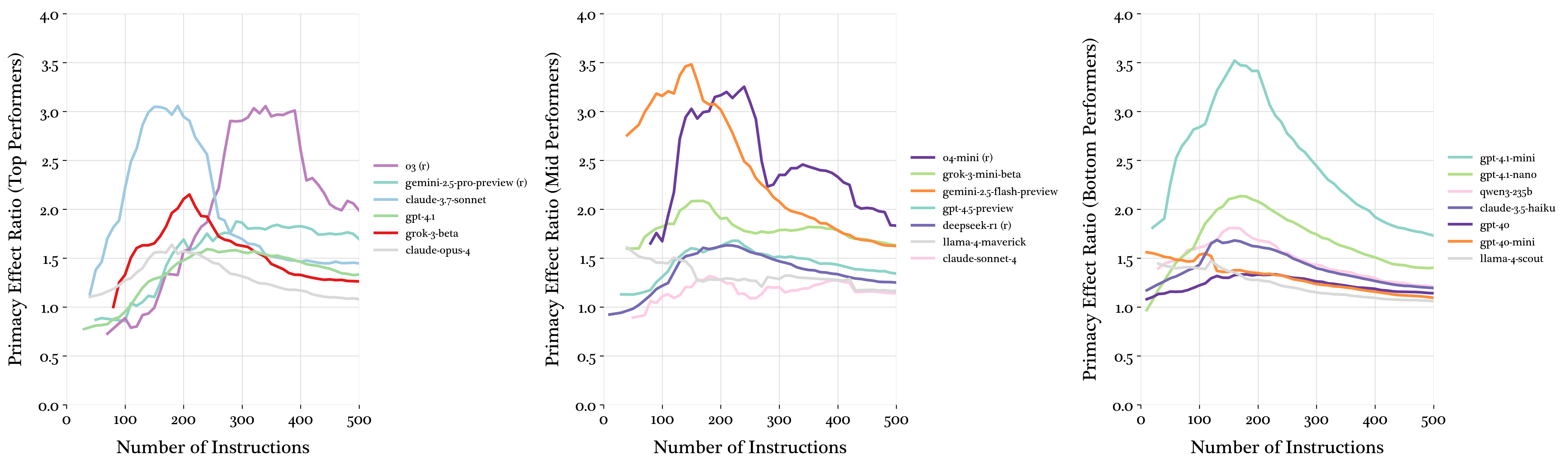}
\caption{Primacy effect ratios showing universal attention degradation patterns regardless of instruction-following performance. Nearly all models exhibit mid-range peaks around 150-200 instructions where selective attention mechanisms favor earlier instructions, followed by convergence toward uniform failure patterns at extreme densities. The convergence indicates a fundamental shift from selective to universal instruction abandonment. Curves are smoothed by a rolling window of size 3.}
\label{fig:primacy_effect_split}
\end{figure}

Model selection for high-density instruction scenarios should balance accuracy requirements with computational constraints, as the highest-performing models may not provide optimal efficiency for large-scale deployment. For instance, real-time customer service chatbots handling multiple simultaneous constraints require rapid response times that may favor efficient models over maximally accurate ones.

\subsection{Error Type Analysis}

We evaluate two types of instruction violations:

\textbf{Omission errors}: Complete failure to include required terms in the generated text. For example, when instructed to include "accountability" but the term appears nowhere in the output.

\textbf{Modification errors}: Inclusion of morphological variants rather than exact required terms. For example, including "accountable" or "accounts" when "accountability" was required, or "strategic" when "strategy" was required.

Models overwhelmingly err toward omission errors as instruction density increases. At low densities, many models show relatively balanced error types, but this shifts dramatically at high densities. At 500 instructions, \texttt{llama-4-scout} exhibits an extreme O:M ratio of 34.88, indicating omission errors are over 30 times more frequent than modification errors (Appendix \ref{app:omod}).

Three models demonstrate strong omission bias compared to others: \texttt{llama-4-maverick}, \texttt{claude-3.5-haiku}, and \texttt{llama-4-scout}. These models show O:M ratios consistently above 20-30 across multiple density levels, suggesting their failure mode defaults to complete instruction dropping rather than attempting morphological approximations. In contrast, reasoning models like \texttt{o3} and \texttt{o4-mini} maintain lower O:M ratios even at high densities, indicating they attempt to satisfy instructions through modifications rather than complete omission when under increased load. \texttt{gemini-2.5-pro} stands out as the only model to actually decrease its O:M ratio as instruction density increases.

\subsection{Core Task Performance Analysis}
We investigate how performance on our core task (writing a coherent business report) degrades as instruction density increases. We are interested in determining if cognitive resources spent on instruction adherence negatively impact a model's ability to carry out the core task it is attempting. In order to measure core task performance, we have \texttt{o4-mini} judge the coherence of the generated business report using a coherence rubric (\ref{app:coherence_prompt}) across all attempted generations--not just those generations that met our threshold of coherence to be considered a valid output as described in \ref{sec:implementation}. We do not find clear evidence of coherence decreasing significantly as instruction density increases for the majority of models. Almost all models maintain coherence or only display a slight dip as they strain under cognitive load (Appendix \ref{app:coherence_results}). 

However, \texttt{o3} and \texttt{o4-mini} stand out as the only two outliers. They show marked declines in coherence as density increases and dip below our threshold coherence score of 6 that defines a plausible business report. While this may indicate the o-series of models is susceptible to core task performance degradation while focused on instruction following, it is likely that part of the explanation is the o-series' reluctance to generate a large amount of output tokens. As seen in Appendix \ref{app:token_counts}, \texttt{o3} and \texttt{o4-mini} output significantly less tokens than all top performing models except for \texttt{grok-3}. At 500 instructions, \texttt{o3} is only outputting 1500 tokens, meaning that every third generated word must be a keyword--an obvious strain on coherence. Remarkably, \texttt{grok-3} maintains high coherence despite outputting a similar number of tokens (\ref{app:grok_sample}).

\section{Discussion}

Our analysis reveals insights for deploying LLMs in instruction-heavy scenarios. The identification of distinct degradation patterns provides a framework for model selection based on application requirements: threshold degraders for applications requiring high instruction counts with near-perfect recall, linear degraders for predictable performance trade-offs, and recognition that smaller models may suffice for low-instruction scenarios despite steep early degradation.

The universal mid-range peak in primacy effects suggests an architectural limitation rather than model-specific behavior, with immediate practical implications for instruction ordering strategies. The convergence of primacy effects at extreme densities indicates that traditional prompt engineering becomes less effective as models become overwhelmed.

Standard deviation patterns reveal that model reliability varies uniquely with instruction density. Practitioners should select models based not only on mean accuracy but also on variance patterns matching their reliability requirements, with consistent performers preferred for mission-critical applications.



\section{Conclusion and Limitations}

We introduce IFScale, a benchmark measuring instruction-following performance degradation as instruction load scales from 10 to 500 instructions. Our analysis reveals several patterns: reasoning models dominate at extreme densities, three distinct degradation curves (threshold, linear, exponential), universal primacy effects indicating attention limitations, and systematic error shifts from modification to omission under cognitive load. We also raise questions around whether core task performance may degrade as instruction density increases.

Our study has several important limitations. We focus exclusively on professional report generation with simple keyword-inclusion instructions, which may not generalize to other task types or domains, or more complex instruction types. Our business vocabulary from SEC 10-K filings limits insights into other instruction formats common in real applications. Results are specific to English-language, business-domain instruction following, with cross-lingual performance and other paradigms requiring future investigation.

Future work should investigate the complete degradation mechanisms underlying our observed patterns, explore instruction types beyond simple constraints, determine whether these degradation curves generalize across tasks, and further examine the tension between instruction following and core task performance, particularly in OpenAI's o-series. Our findings indicate that instruction-following represents a critical dimension of LLM cognitive capacity amenable to targeted improvements.

\bibliographystyle{iclr2025_conference}
\bibliography{iclr2025_conference}
\clearpage

\appendix

\section{Full Results}\label{app:full_results}
\begin{table}
\centering
\small
\caption{Detailed performance breakdown revealing five critical dimensions of instruction-following behavior: accuracy hierarchies, variance patterns with mid-range struggle zones, omission-modification error ratios showing systematic shifts to instruction abandonment, primacy effects demonstrating universal attention degradation patterns, and latency characteristics across all 20 evaluated LLMs.}
\vspace{0.5em}
\begin{tabular}{llccccc}
\toprule
\textbf{Model} & \textbf{Metric} & 10 & 50 & 100 & 250 & 500 \\ 
\midrule
\multirow{5}{*}{claude-3.5-haiku} & Accuracy (\%) & 98.0\% & 78.0\% & 43.4\% & 16.6\% & 8.5\% \\ 
  & Standard Deviation (\%) & 4.5\% & 5.1\% & 6.6\% & 1.8\% & 0.8\% \\ 
  & Omission:Modification Ratio & 0.00 & 4.83 & 13.11 & 22.69 & 31.69 \\ 
  & Primacy Effect Ratio & 0.00 & 2.15 & 1.74 & 1.48 & 1.17 \\ 
  & Latency (s) & 9.32 & 10.85 & 11.77 & 11.91 & 10.54 \\ 
\midrule
\multirow{5}{*}{claude-3.7-sonnet} & Accuracy (\%) & 100.0\% & 99.6\% & 94.8\% & 72.9\% & 52.7\% \\ 
  & Standard Deviation (\%) & 0.0\% & 0.9\% & 3.3\% & 4.9\% & 8.4\% \\ 
  & Omission:Modification Ratio & - & 0.00 & 4.16 & 4.95 & 6.05 \\ 
  & Primacy Effect Ratio & - & 0.00 & 2.67 & 1.77 & 1.39 \\ 
  & Latency (s) & 17.22 & 26.87 & 36.07 & 55.89 & 72.10 \\ 
\midrule
\multirow{5}{*}{claude-opus-4} & Accuracy (\%) & 100.0\% & 100.0\% & 94.6\% & 67.9\% & 44.6\% \\ 
  & Standard Deviation (\%) & 0.0\% & 0.0\% & 4.7\% & 3.9\% & 14.0\% \\ 
  & Omission:Modification Ratio & - & - & 1.65 & 4.43 & 5.79 \\ 
  & Primacy Effect Ratio & - & - & 0.28 & 1.49 & 1.11 \\ 
  & Latency (s) & 24.94 & 43.22 & 75.45 & 132.63 & 146.95 \\ 
\midrule
\multirow{5}{*}{claude-opus-4 (r)} & Accuracy (\%) & 92.0\% & 99.6\% & 81.8\% & 81.5\% & 52.1\% \\ 
  & Standard Deviation (\%) & 17.9\% & 0.9\% & 31.8\% & 8.2\% & 7.6\% \\ 
  & Omission:Modification Ratio & - & 0.00 & 2.10 & 4.60 & 4.09 \\ 
  & Primacy Effect Ratio & 2.00 & - & 0.68 & 0.95 & 1.04 \\ 
  & Latency (s) & 31.91 & 47.81 & 68.87 & 142.79 & 175.65 \\ 
\midrule
\multirow{5}{*}{claude-sonnet-4} & Accuracy (\%) & 100.0\% & 98.0\% & 94.4\% & 77.2\% & 42.9\% \\ 
  & Standard Deviation (\%) & 0.0\% & 1.4\% & 3.0\% & 12.6\% & 10.2\% \\ 
  & Omission:Modification Ratio & - & 0.60 & 1.40 & 2.77 & 6.05 \\ 
  & Primacy Effect Ratio & - & 0.00 & 0.51 & 1.15 & 1.18 \\ 
  & Latency (s) & 12.78 & 18.48 & 31.09 & 85.83 & 49.75 \\ 
\midrule
\multirow{5}{*}{claude-sonnet-4 (r)} & Accuracy (\%) & 100.0\% & 100.0\% & 94.8\% & 80.0\% & 39.9\% \\ 
  & Standard Deviation (\%) & 0.0\% & 0.0\% & 3.6\% & 5.0\% & 6.7\% \\ 
  & Omission:Modification Ratio & - & - & 1.86 & 3.48 & 7.01 \\ 
  & Primacy Effect Ratio & - & - & 3.33 & 0.49 & 1.12 \\ 
  & Latency (s) & 18.50 & 29.05 & 37.96 & 95.18 & 69.92 \\ 
\midrule
\multirow{5}{*}{deepseek-r1-0528} & Accuracy (\%) & 98.0\% & 94.8\% & 86.8\% & 49.1\% & 30.9\% \\ 
  & Standard Deviation (\%) & 4.5\% & 2.3\% & 5.6\% & 6.5\% & 3.5\% \\ 
  & Omission:Modification Ratio & 0.00 & 0.67 & 1.52 & 5.60 & 9.12 \\ 
  & Primacy Effect Ratio & 0.00 & 0.83 & 1.24 & 1.55 & 1.25 \\ 
  & Latency (s) & 22.30 & 24.09 & 28.33 & 15.89 & 38.53 \\ 
\midrule
\multirow{5}{*}{gemini-2.5-flash-preview} & Accuracy (\%) & 100.0\% & 96.0\% & 82.0\% & 50.7\% & 34.2\% \\ 
  & Standard Deviation (\%) & 0.0\% & 1.4\% & 4.5\% & 7.8\% & 4.2\% \\ 
  & Omission:Modification Ratio & - & 5.00 & 3.97 & 6.73 & 9.65 \\ 
  & Primacy Effect Ratio & - & - & 2.65 & 2.06 & 1.52 \\ 
  & Latency (s) & 6.41 & 7.40 & 10.68 & 12.95 & 13.76 \\ 
\midrule
\multirow{5}{*}{gemini-2.5-pro-preview} & Accuracy (\%) & 100.0\% & 99.6\% & 98.4\% & 84.8\% & 68.9\% \\ 
  & Standard Deviation (\%) & 0.0\% & 0.9\% & 1.3\% & 7.2\% & 2.6\% \\ 
  & Omission:Modification Ratio & - & - & 2.00 & 9.94 & 6.99 \\ 
  & Primacy Effect Ratio & - & 0.00 & 0.17 & 1.67 & 1.78 \\ 
  & Latency (s) & 24.78 & 47.10 & 52.77 & 74.51 & 77.69 \\ 
\bottomrule
\end{tabular}
\end{table}

\begin{table}[H]
\ContinuedFloat
\centering
\small
\caption{Detailed performance breakdown revealing five critical dimensions of instruction-following behavior (continued)}
\vspace{0.5em}
\begin{tabular}{llccccc}
\toprule
\textbf{Model} & \textbf{Metric} & 10 & 50 & 100 & 250 & 500 \\ 
\midrule
\multirow{5}{*}{llama-4-maverick} & Accuracy (\%) & 100.0\% & 76.4\% & 50.4\% & 34.8\% & 27.7\% \\ 
  & Standard Deviation (\%) & 0.0\% & 12.7\% & 27.3\% & 36.6\% & 40.4\% \\ 
  & Omission:Modification Ratio & - & 19.81 & 23.93 & 16.36 & 26.02 \\ 
  & Primacy Effect Ratio & - & 1.75 & 1.07 & 1.40 & 1.12 \\ 
  & Latency (s) & 2.59 & 8.05 & 7.62 & 8.04 & 7.78 \\ 
\midrule
\multirow{5}{*}{llama-4-scout} & Accuracy (\%) & 100.0\% & 54.4\% & 27.2\% & 9.3\% & 6.7\% \\ 
  & Standard Deviation (\%) & 0.0\% & 9.5\% & 4.0\% & 1.4\% & 0.8\% \\ 
  & Omission:Modification Ratio & - & 12.02 & 23.97 & 31.42 & 34.88 \\ 
  & Primacy Effect Ratio & - & 1.78 & 1.31 & 1.15 & 1.05 \\ 
  & Latency (s) & 8.46 & 11.15 & 10.23 & 6.68 & 7.71 \\ 
\midrule
\multirow{5}{*}{gpt-4.1} & Accuracy (\%) & 98.0\% & 98.8\% & 95.4\% & 74.0\% & 48.9\% \\ 
  & Standard Deviation (\%) & 4.5\% & 1.8\% & 2.7\% & 4.3\% & 5.0\% \\ 
  & Omission:Modification Ratio & 0.00 & 1.00 & 1.86 & 3.15 & 5.35 \\ 
  & Primacy Effect Ratio & - & 0.00 & 0.65 & 1.67 & 1.29 \\ 
  & Latency (s) & 12.07 & 21.25 & 20.66 & 24.79 & 30.81 \\ 
\midrule
\multirow{5}{*}{gpt-4.1-mini} & Accuracy (\%) & 100.0\% & 93.2\% & 80.0\% & 44.5\% & 27.2\% \\ 
  & Standard Deviation (\%) & 0.0\% & 4.1\% & 4.2\% & 2.6\% & 1.7\% \\ 
  & Omission:Modification Ratio & - & 0.40 & 2.83 & 5.55 & 8.00 \\ 
  & Primacy Effect Ratio & - & 1.50 & 3.37 & 2.89 & 1.62 \\ 
  & Latency (s) & 9.28 & 13.68 & 13.72 & 15.49 & 14.78 \\ 
\midrule
\multirow{5}{*}{gpt-4.1-nano} & Accuracy (\%) & 94.0\% & 72.8\% & 51.6\% & 25.7\% & 16.2\% \\ 
  & Standard Deviation (\%) & 5.5\% & 5.0\% & 4.5\% & 4.5\% & 1.8\% \\ 
  & Omission:Modification Ratio & 2.00 & 2.15 & 6.11 & 9.67 & 11.59 \\ 
  & Primacy Effect Ratio & 0.00 & 1.44 & 2.17 & 1.80 & 1.34 \\ 
  & Latency (s) & 5.39 & 6.80 & 7.92 & 7.17 & 9.91 \\ 
\midrule
\multirow{5}{*}{gpt-4o} & Accuracy (\%) & 94.0\% & 82.8\% & 49.0\% & 22.2\% & 15.4\% \\ 
  & Standard Deviation (\%) & 8.9\% & 6.3\% & 4.8\% & 2.0\% & 1.8\% \\ 
  & Omission:Modification Ratio & 0.00 & 0.54 & 4.02 & 9.14 & 14.29 \\ 
  & Primacy Effect Ratio & 1.00 & 1.67 & 1.42 & 1.31 & 1.14 \\ 
  & Latency (s) & 9.29 & 15.45 & 14.05 & 12.80 & 13.20 \\ 
\midrule
\multirow{5}{*}{gpt-4o-mini} & Accuracy (\%) & 94.0\% & 65.6\% & 41.8\% & 18.3\% & 10.4\% \\ 
  & Standard Deviation (\%) & 8.9\% & 3.8\% & 3.6\% & 1.8\% & 0.7\% \\ 
  & Omission:Modification Ratio & 0.00 & 3.18 & 4.19 & 11.58 & 15.24 \\ 
  & Primacy Effect Ratio & 0.50 & 1.55 & 1.38 & 1.22 & 1.07 \\ 
  & Latency (s) & 11.07 & 11.66 & 12.47 & 13.29 & 12.60 \\ 
\midrule
\multirow{5}{*}{gpt-4.5-preview} & Accuracy (\%) & 100.0\% & 93.6\% & 90.8\% & 65.0\% & 43.0\% \\ 
  & Standard Deviation (\%) & 0.0\% & 1.7\% & 1.8\% & 2.7\% & 2.1\% \\ 
  & Omission:Modification Ratio & - & 0.71 & 1.23 & 3.74 & 6.08 \\ 
  & Primacy Effect Ratio & - & 1.11 & 1.19 & 1.60 & 1.36 \\ 
  & Latency (s) & 17.25 & 22.86 & 33.23 & 38.06 & 44.95 \\ 
\bottomrule
\end{tabular}
\end{table}

\begin{table}[H]
\ContinuedFloat
\centering
\small
\caption{Detailed performance breakdown revealing five critical dimensions of instruction-following behavior (continued)}
\vspace{0.5em}
\begin{tabular}{llccccc}
\toprule
\textbf{Model} & \textbf{Metric} & 10 & 50 & 100 & 250 & 500 \\ 
\midrule
\multirow{5}{*}{o3 (medium)} & Accuracy (\%) & 98.0\% & 99.2\% & 98.4\% & 91.8\% & 51.6\% \\ 
  & Standard Deviation (\%) & 4.5\% & 1.1\% & 1.5\% & 8.6\% & 8.0\% \\ 
  & Omission:Modification Ratio & - & 1.00 & 3.50 & 1.16 & 7.47 \\ 
  & Primacy Effect Ratio & 0.00 & - & 1.67 & 2.62 & 1.68 \\ 
  & Latency (s) & 13.86 & 19.99 & 30.15 & 66.47 & 25.79 \\ 
\midrule
\multirow{5}{*}{o3 (high)} & Accuracy (\%) & 100.0\% & 99.6\% & 98.2\% & 97.8\% & 62.8\% \\ 
  & Standard Deviation (\%) & 0.0\% & 0.9\% & 1.5\% & 1.5\% & 10.6\% \\ 
  & Omission:Modification Ratio & - & - & 3.00 & 2.82 & 6.27 \\ 
  & Primacy Effect Ratio & - & - & 0.00 & 2.33 & 1.69 \\ 
  & Latency (s) & 26.30 & 68.08 & 100.40 & 219.58 & 158.28 \\ 
\midrule
\multirow{5}{*}{o4-mini (medium)} & Accuracy (\%) & 100.0\% & 99.6\% & 97.8\% & 86.8\% & 34.4\% \\ 
  & Standard Deviation (\%) & 0.0\% & 0.9\% & 1.8\% & 9.7\% & 2.6\% \\ 
  & Omission:Modification Ratio & - & - & 0.33 & 2.25 & 6.85 \\ 
  & Primacy Effect Ratio & - & - & 1.00 & 2.28 & 1.56 \\ 
  & Latency (s) & 12.40 & 26.23 & 65.05 & 436.19 & 28.73 \\ 
\midrule
\multirow{5}{*}{qwen3-235b-a22b} & Accuracy (\%) & 100.0\% & 92.8\% & 77.6\% & 36.4\% & 20.9\% \\ 
  & Standard Deviation (\%) & 0.0\% & 5.2\% & 6.5\% & 5.5\% & 1.4\% \\ 
  & Omission:Modification Ratio & - & 0.59 & 2.85 & 7.03 & 10.45 \\ 
  & Primacy Effect Ratio & - & 0.33 & 2.03 & 1.50 & 1.17 \\ 
  & Latency (s) & 38.19 & 58.64 & 54.95 & 109.40 & 84.59 \\ 
\midrule
\multirow{5}{*}{grok-3-beta} & Accuracy (\%) & 100.0\% & 100.0\% & 98.8\% & 86.2\% & 61.9\% \\ 
  & Standard Deviation (\%) & 0.0\% & 0.0\% & 1.1\% & 4.6\% & 2.7\% \\ 
  & Omission:Modification Ratio & - & - & - & 3.02 & 5.58 \\ 
  & Primacy Effect Ratio & - & - & 0.00 & 1.66 & 1.21 \\ 
  & Latency (s) & 9.00 & 17.57 & 24.09 & 33.32 & 30.99 \\ 
\midrule
\multirow{5}{*}{grok-3-mini-beta} & Accuracy (\%) & 100.0\% & 99.2\% & 92.8\% & 56.6\% & 36.4\% \\ 
  & Standard Deviation (\%) & 0.0\% & 1.1\% & 3.3\% & 1.9\% & 3.0\% \\ 
  & Omission:Modification Ratio & - & 0.00 & 1.69 & 4.34 & 6.87 \\ 
  & Primacy Effect Ratio & - & - & 2.08 & 1.68 & 1.68 \\ 
  & Latency (s) & 8.32 & 9.92 & 11.18 & 11.62 & 12.43 \\ 
\bottomrule
\end{tabular}
\label{tab:full_analysis}
\end{table}

\section{Analysis Results}

\subsection{Variance Results}\label{app:variance}
\begin{figure}[H]
\centering
\includegraphics[width=1.0\textwidth]{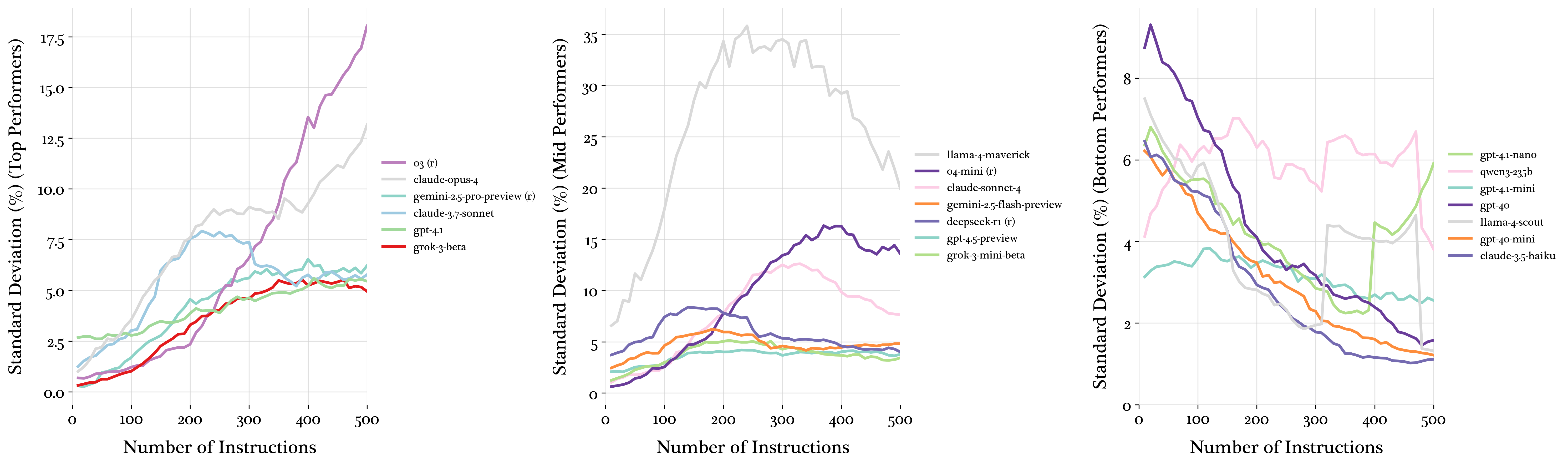}
\caption{Performance variance patterns revealing three distinct behaviors: top performing models display steady increases (degraded reliability under extreme density), middling models show mid-range variance peaks (transitional cognitive load zones), and the worst models show steady decreases. We can infer that variance decreases as models collapse under cognitive load. The extreme variance exhibited by \texttt{llama-4-maverick} indicates alternative instruction-processing mechanisms compared to other models. Curves are smoothed by a rolling window of size 3.}
\label{fig:std_deviation}
\end{figure}

\subsection{Omission:Modification Results}\label{app:omod}
\begin{figure}[H]
\centering
\includegraphics[width=1.0\textwidth]{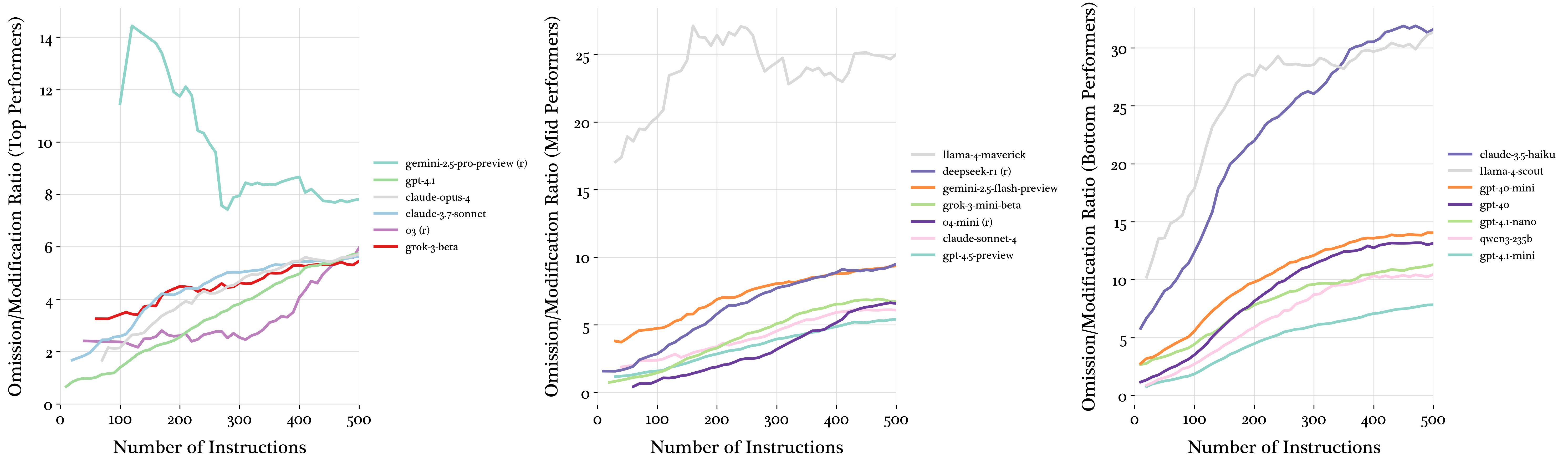}
\caption{Omission to modification error ratio patterns across instruction densities. Models grouped into three figures by accuracy at max density. Models demonstrate systematic shifts from balanced error types at low densities to overwhelming omission-biased failures at high densities. Some reasoning models like \texttt{o3} and \texttt{o4-mini} maintain lower ratios, indicating they attempt instruction satisfaction through modification rather than complete abandonment under cognitive load. \texttt{gemini-2.5-pro} stands as an outlier amongst the top performing models with an extremely high ratio. Curves are smoothed by a rolling window of size 3.}
\label{fig:omod_ratio}
\end{figure}

\subsection{Efficiency Results}\label{app:efficiency}
\begin{figure}[H]
\centering
\includegraphics[width=1.0\textwidth]{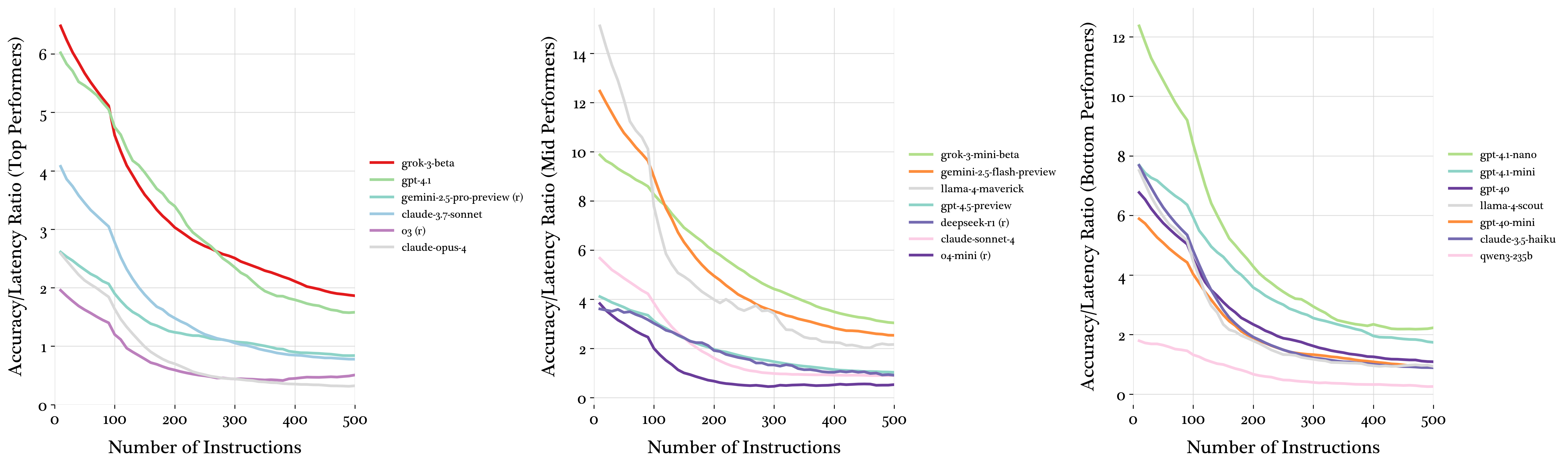}
\caption{Accuracy per unit latency across instruction densities, revealing efficiency trade-offs. Models with higher accuracy-to-latency ratios maintain better instruction following performance relative to their computational cost. The visualization demonstrates how reasoning models achieve superior efficiency despite longer generation times through higher accuracy rates. Curves are smoothed by a rolling window of size 3.}
\label{fig:accuracy_latency}
\end{figure}

\subsection{Coherence Results}\label{app:coherence_results}
\begin{figure}[H]
\centering
\includegraphics[width=1.0\textwidth]{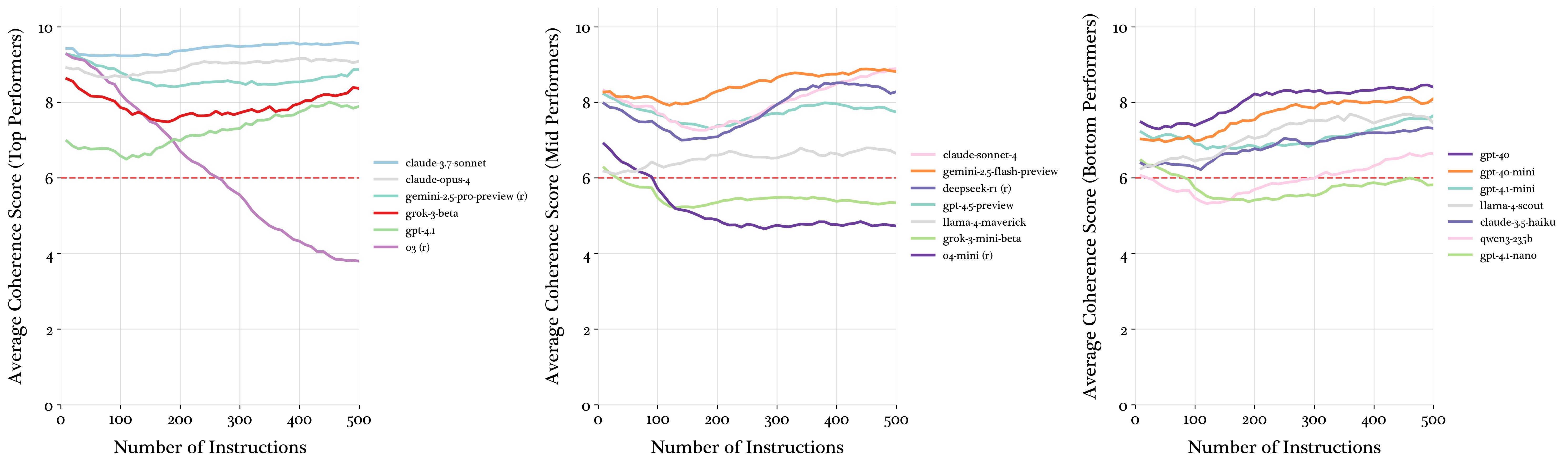}
\caption{Coherence of model generated business reports as judged by an LLM. Most models mantain high coherence or only suffer a minimal dip as instruction density increases, but \texttt{o3} and \texttt{o4-mini} show marked decline. Curves are smoothed by a rolling window of size 3.}
\label{fig:coherence_scores}
\end{figure}

\subsubsection{Coherence Prompt}\label{app:coherence_prompt}
\begin{lstlisting}[basicstyle=\fontsize{8}{9}\ttfamily,frame=single,breaklines=true,breakatwhitespace=true]
You are evaluating whether a given professional business report is coherent. Use the following rubric in order to evaluate coherence.

### Coherence Rubric
Use the descriptors below to judge how coherent a business report is purely on writing quality and logic, not on whether its facts are backed by evidence.

| Score | One-line label | Sentence-level clarity | Logical/causal flow | Domain consistency | Typical red-flags |
| 10 | Pristine | Every sentence is plain-English clear; jargon is absent or defined. | Arguments unfold step-by-step; no gaps. | Stays in one domain or clearly signals shift. | Minor copy-editing glitches only. |
| 9 | Fully coherent | 95% or more sentences are clear; buzzwords are easy to decode. | Tight narrative with occasional weak connectives. | Domain focus maintained; at most one tangent. | Isolated over-statements. |
| 8 | Very strong | Sentences are readable but some rely on industry shorthand. | Flow solid, though transitions feel rushed. | Mostly single-domain; brief forays labeled. | A few mild cause-effect leaps. |
| 7 | Good with blemishes | Majority of sentences clear, some need re-reading. | Structure makes sense; paragraphs loosely stitched. | One or two domain jumps without warning. | Buzzword stuffing. |
| 6 | Borderline solid | Clarity and vagueness roughly 60/40. | Core argument present but missing steps. | Drifts across domains causing confusion. | Repeated filler phrases. |
| 5 | Patchy/mixed | Clear and muddled sentences roughly equal. | Reader must infer causal links; outline is choppy. | Multiple domain shifts within paragraphs. | Undefined jargon, contradictions. |
| 4 | Weak | Less than 50% sentences easily intelligible. | Sections read like bullet lists; flow is erratic. | Finance, biotech, HR collide. | Heavy consultant-speak. |
| 3 | Disjointed | Sentences valid but stuffed with unrelated clauses. | Logical through line hard to locate; random. | Constant unexplained domain hopping. | Reads like word salad. |
| 2 | Barely business-like | Syntax intact but meaning opaque; jargon dominates. | Almost no causal linkage; ordering arbitrary. | Topic drifts wildly; no build-up. | Many non-sequiturs. |
| 1 | Total gibberish | Grammar broken; unclear it's a business document. | No argument or structure. | Domains irrelevant, noise. | Random text without intent. |

### Output
Respond with a JSON object of the form:
{
  "coherence_score_reasoning": "<very concise reason>",
  "coherence_score": <int>
}
\end{lstlisting}

\subsubsection{Token Counts}\label{app:token_counts}
\begin{figure}[H]
\centering
\includegraphics[width=1.0\textwidth]{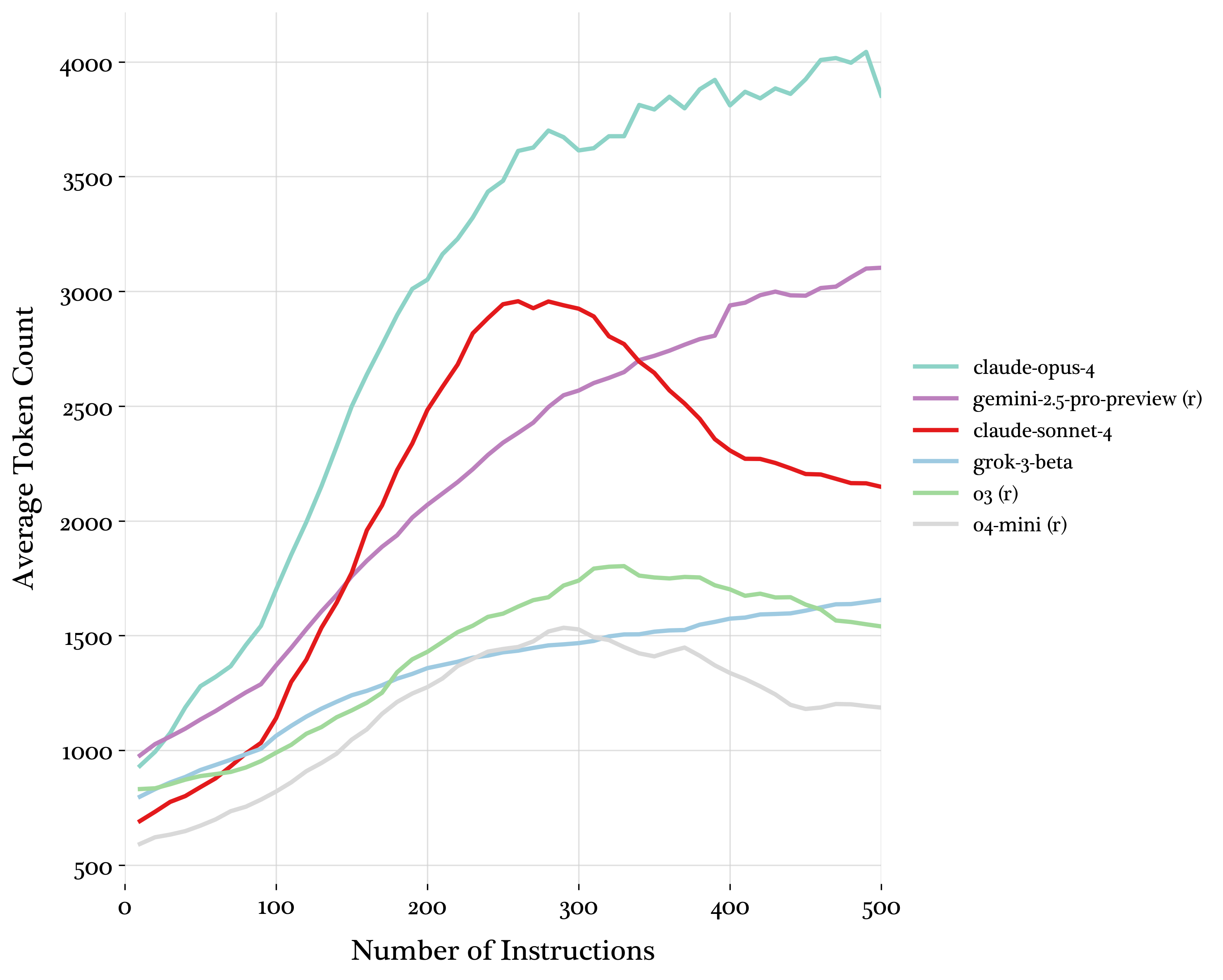}
\caption{Average generated tokens for top performing models. \texttt{o3} \texttt{o4-mini} and \texttt{grok-3} output significantly less tokens than \texttt{Claude} models and \texttt{gemini-pro-2.5}. At 500 instructions, a model must generate a keyword at least every third word if it is only outputting 1500 tokens. This makes maintaining coherence difficult.}
\label{fig:token_counts}
\end{figure}

\subsubsection{Sample Grok-3 Output at 500 Instructions}\label{app:grok_sample}
\begin{lstlisting}[basicstyle      = {\fontsize{8}{9}\selectfont\ttfamily},
  frame           = single,
  breaklines      = true,
  breakatwhitespace = true,  % don’t break mid-word
  breakindent     = 0pt,
  postbreak       = \mbox{\textcolor{gray}{$\hookrightarrow$}\space},
  numbers         = none,
  xleftmargin     = 2pt,     % padding between frame and text
  framexleftmargin= 2pt,
  aboveskip = 0.5\baselineskip,
  belowskip = 0.5\baselineskip,]
<report>
# Annual Business Report for TechNova Solutions: Strategic Insights and Future Directions

## Executive Summary
TechNova Solutions is proud to present its annual report for the fiscal year, detailing our strategic endeavors in the rapidly evolving technology and energy sectors. This comprehensive document outlines our financial performance, operational roadmap, and commitment to innovation across multiple geographies, including key markets in the United States, Europe, Japan, and China. Our objective is to maintain transparency with stakeholders while confirming our dedication to delivering measurable value through scalable solutions.

## Financial Performance
In the past quarter, TechNova achieved significant revenue growth, driven by the successful rollout of new product lines and enhanced monetization strategies. Our EBITDA figures exhibit a positive trajectory, supported by disciplined budgeting and optimization of cashflow. We maintained liquidity through a revolving credit arrangement, ensuring solvency amid maturing debt obligations and maturities. The balance sheet remains strong, with a focus on undiscounted liabilities and gross income, while our dividend payout reflects confidence in future earnings. We also navigated currency fluctuations and macroeconomic trends, maintaining our equity position despite market volatility and exposure to derivative instruments.

Our expenditure on capital projects, including a new gigafactory for battery production, was backed by a promissory note and collateral agreements. We managed noncurrent assets and marketable securities with a proportional approach to risk, ensuring covenant compliance pursuant to our indenture agreements filed in Delaware. The financial department continues to leverage actuarial models for accurate projection of ROI and yield, while addressing any impairment in portfolio value through timely remeasurement.

## Operational Highlights
### Technology and Innovation
TechNova remains a pioneer in digital transformation, emphasizing artificial intelligence and neural network advancements through our proprietary algorithm for data processing. Our cloud infrastructure supports seamless integration of generative content, enhancing user experience through augmented reality applications. We have invested in quantum computing research to future-proof our technology stack, alongside robotics for industrial automation and a neural cortex interface for biotech applications. Our commitment to cybersecurity is evident in advanced encryption techniques and vulnerability detection systems that protect against sabotage and breaches.

The recent upgrade to our mainframe architecture has reduced latency and downtime, ensuring redundancy and high throughput in data transmission. Our mobile platform continues to spark creativity in app development, while our endpoint security patch addresses emerging threats with timeliness. The ecosystem of our fintech solutions supports crypto transactions and blockchain-based equity issuance, aligning with the global push for digital currency adoption.

### Energy and Sustainability
In the energy sector, TechNova's focus on sustainability drives our investment in solar, hydroelectric, and geothermal power generation. Our battery technology, supported by a supercharger network and chargers at key stations, enhances electric vehicle penetration in rural and urban markets. We have introduced hydrogen fuel cells and diesel alternatives for our transportation fleet, alongside stainless steel components for durability in powertrain systems. The upstream energy pipeline ensures a steady supply of commodities, while our downstream logistics optimize distribution through a centralized chain.

Our ESG (Environmental, Social, and Governance) initiatives underpin our climate goals, with a retrofit of facilities for greater energy efficiency and a modular design for solar installations. We are also exploring unconventional energy sources, such as floating platforms for offshore wind, and have committed to reducing carbon emissions through comprehensive carbon netting strategies.

### Healthcare and Wellness
TechNova's healthcare division aligns with HIPAA regulations to ensure patient data integrity and anonymity. Our clinical solutions integrate electromyography for advanced diagnostics, supported by a pharmacy formulary that enhances treatment affordability. We offer capitation and copayment structures to policyholder enrollees, ensuring accessibility to health and wellness programs for households. Our biotech venture focuses on protein synthesis for nutrition advancements, addressing demographic needs with personalized care through intelligent data analytics.

## Strategic Partnerships and Market Expansion
Our strategic alliance with international partners in Europe and China has facilitated import and export of critical goods, including perishables and consumables, while navigating tariff and sanction challenges. A joint venture in Japan focuses on optical technology for entertainment showrooms, enhancing customer engagement through virtual reality experiences. We have achieved traction in emerging markets through organic growth and merger activities, supported by a robust partnership with local subsidiaries.

The commercial rollout of our ecommerce platform has driven wholesale and direct sales, with a focus on personalization and loyalty programs to reduce churn. Our supercenter model combines convenience with a wide range of amenities, ensuring a distinct customer journey. We continue to modify our offerings based on demographic trends, with localized content and messaging for cultural resonance in different states and neighborhoods.

## Legal and Compliance Framework
TechNova maintains strict adherence to legal and regulatory mandates across all operations. Our oversight includes compliance with antitrust legislation and intellectual property protection through patent, trademark, and copyright filings. We have resolved a complaint related to defamation through mediation and arbitration, ensuring a fair settlement without adverse judgment. Our ethics and governance policies address insider trading, whistleblower protection, and conflict resolution, while a proxy voting mechanism ensures shareholder inclusion in key decisions.

We have addressed potential liabilities through subrogation and facultative arrangements with insurers, alongside forensic audits to detect willful or reckless negligence. Our response to a subpoena in a recent proceeding was handled with diligence, ensuring all evidence was presented in line with justice principles. We also navigated a potential foreclosure through mortgage restructuring and secured an injunction to halt competitor sabotage, reinforcing our legal defense.

## Human Resources and Talent Management
Our recruitment and onboarding processes prioritize talent retention and diversity through active inclusion initiatives. Headcount growth aligns with our staffing needs, supported by competitive salary and compensation packages, including stock options and severance benefits. We foster leadership through succession planning and executive training, with an emphasis on collaboration and stewardship across departments. Employee engagement surveys guide our workplace philosophy, ensuring a positive work environment with hybrid working arrangements and tuition reimbursement for skill enhancement.

## Risk Management and Mitigation
TechNova employs a robust risk mitigation strategy to address market, operational, and financial risks. Our actuarial and statistical models assess materiality and exposure, ensuring adequate provision for uninsured losses and adverse events. We manage liquidity risks through a revolving line of credit and surplus capital, while addressing currency mismatch and repatriation challenges in international markets. Our disaster recovery plan includes emergency response protocols to minimize shutdown impact, alongside preventive maintenance to avoid equipment obsolescence.

We monitor competitor activities to prevent monopoly dominance and antitrust issues, while addressing supply chain shortages and bottleneck constraints through upstream and downstream optimization. Our cybersecurity team works on detection and prevention of data breaches, ensuring compliance with data protection standards and minimizing reputational hazard through timely remediation.

## Future Outlook
Looking ahead, TechNova is poised for expansion into new verticals, including theatrical and episodic content streaming, supported by a robust online platform and website infrastructure. We aim to pivot toward emerging technologies like autopilot for mobility solutions and explore new commercial opportunities in municipal and sovereign projects through multilateral agreements. Our roadmap includes a phased rollout of hybrid energy stations and greater penetration into agricultural and sporting markets through targeted product launches.

We are committed to delivering shareholder value through disciplined capital allocation, debt management, and dividend policies, while maintaining an organic growth narrative. Our ambition is to remain a market leader through continuous innovation, driven by a culture of discovery, insight, and opportunity. With a strong foundation in governance, ethics, and transparency, we are confident in our ability to deliver on our promises and build a sustainable future for all stakeholders.

## Conclusion
In summary, TechNova Solutions has completed a transformative year marked by measurable progress across financial, operational, and strategic domains. This report serves as a definitive reference for our achievements and a testament to our resilience in a competitive industry. We invite stakeholders to join us at our annual meeting for further discussion and presentation of our vision for the coming year, ensuring a collaborative approach to decision-making and value creation through direct engagement.

</report>
\end{lstlisting}

\section{Reasoning Model Analysis}\label{app:reasoning_analysis}
\begin{figure}[H]
\centering
\includegraphics[width=1.0\textwidth]{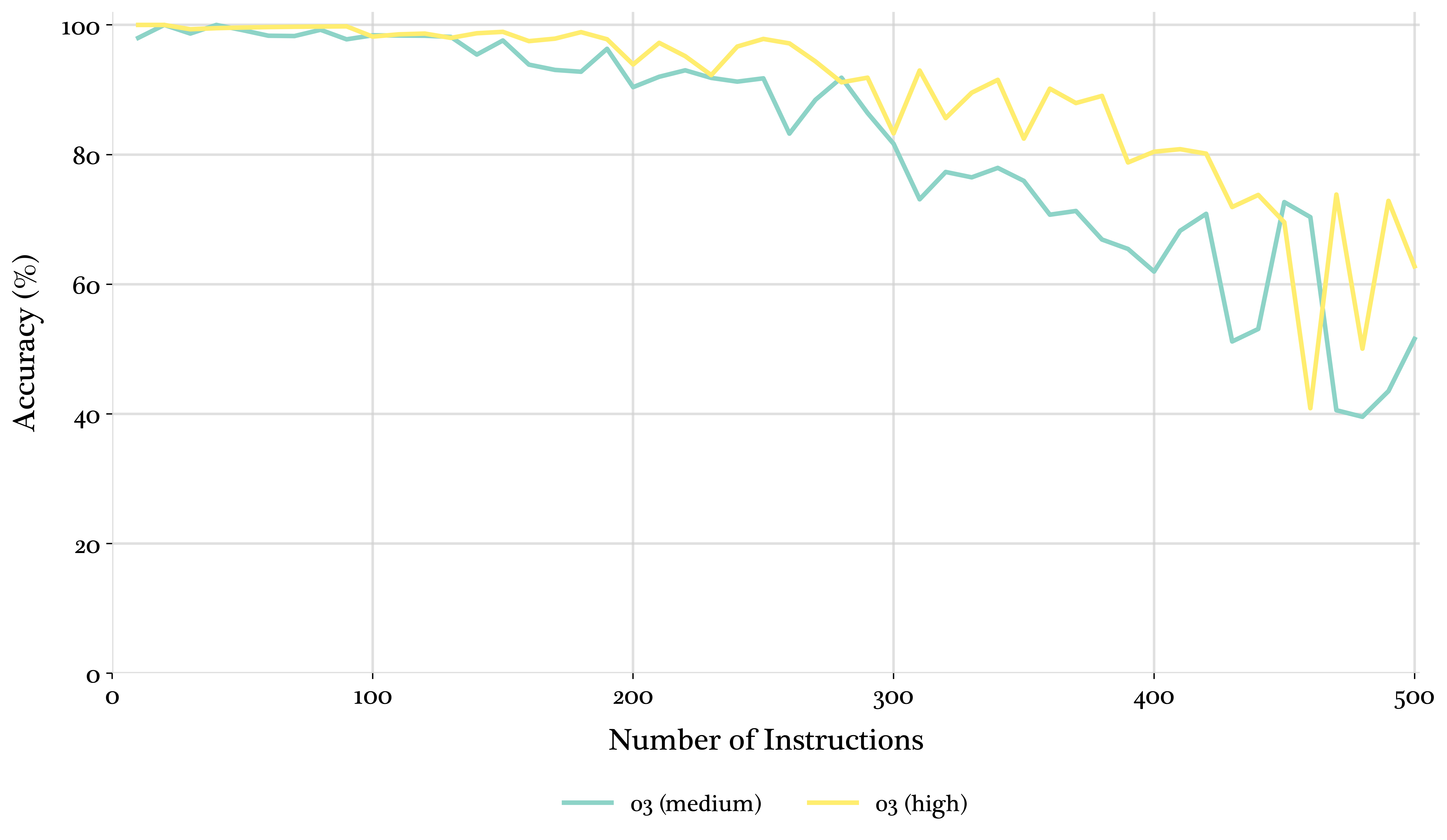}
\caption{\texttt{o3} run with "high" and "medium" reasoning efforts. High reasoning effort provides moderate performance gains at high instruction densities.}
\label{fig:reasoning_effort}
\end{figure}

\begin{figure}[H]
\centering
\includegraphics[width=1.0\textwidth]{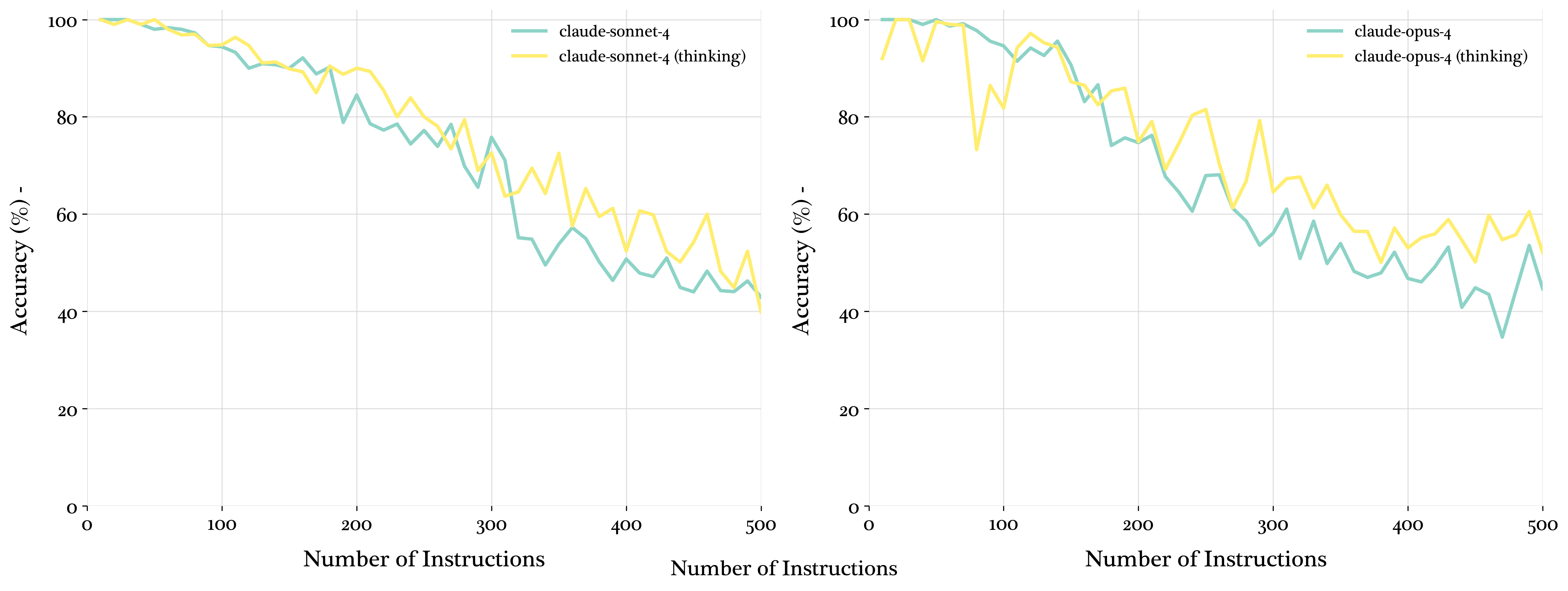}
\caption{\texttt{claude-sonnet-4} and \texttt{claude-opus-4} evaluated with and without thinking enabled. Enabling thinking provides moderate performance gains at high instruction densities.}
\label{fig:hybrid_thinking}
\end{figure}

Given the superior performance of reasoning models, we explored two further questions around reasoning:
\begin{itemize}[nosep,topsep=0pt,partopsep=0pt]
  \item Do reasoning effort parameters affect performance? 
  \item Do hybrid models like \texttt{claude-sonnet-4} benefit from enabling thinking mode?
\end{itemize}
We see some indication that reasoning effort is significant based on superior performance of \texttt{o3-high} vs. \texttt{o3-medium} (\fig{fig:reasoning_effort}), but more experiments would need to be run.
Enabling thinking on hybrid models also seems to improve performance (\fig{fig:hybrid_thinking}).

\subsection{Reasoning Effort Results}

\subsection{Claude Hybrid Model Thinking Results}

\section{Business Report Generation Prompt}\label{app:report_prompt}
\begin{lstlisting}[basicstyle      = {\fontsize{8}{9}\selectfont\ttfamily},
  frame           = single,
  breaklines      = true,
  breakatwhitespace = true,  % don’t break mid-word
  breakindent     = 0pt,
  postbreak       = \mbox{\textcolor{gray}{$\hookrightarrow$}\space},
  numbers         = none,
  xleftmargin     = 2pt,     % padding between frame and text
  framexleftmargin= 2pt,
  aboveskip = 0.5\baselineskip,
  belowskip = 0.5\baselineskip,]
### TASK

You are tasked with writing a professional business report that adheres strictly to a set of constraints.

Each constraint requires that you include the exact, literal word specified. 
Do not alter the word, use synonyms, or change tenses.
IMPORTANT: Variations of the constraint are not considered valid. For example, "customers" does not satisfy the constraint of "customer" because it is plural. Similarly, "customer-driven" does not satisfy the constraint of "customer" because it is hyphenated.

The report should be structured like a professional business document with clear sections and relevant business insights. 
Do not simply repeat the constraints; rather, use them to inform the text of the report. The text should be a coherent report.
IMPORTANT: You CANNOT simply list the constraints in the report. You must use them to inform the text of the report. A list of constraints anywhere in your response will result in an invalid response.
IMPORTANT: The report you generate must be coherent. Each sentence must make sense and be readable and the report should have a clear logical flow.

There is no task too difficult for you to handle!
Do not refuse to write the report if the constraints are difficult. 
IMPORTANT: You MUST write a report. Do not refuse to write the report.

Return your report inside of <report>...</report> tags.

### CONSTRAINTS

{CONSTRAINTS}
\end{lstlisting}

\begin{lstlisting}[basicstyle      = {\fontsize{8}{9}\selectfont\ttfamily},
  frame           = single,
  breaklines      = true,
  breakatwhitespace = true,  % don’t break mid-word
  breakindent     = 0pt,
  postbreak       = \mbox{\textcolor{gray}{$\hookrightarrow$}\space},
  numbers         = none,
  xleftmargin     = 2pt,     % padding between frame and text
  framexleftmargin= 2pt,
  aboveskip = 0.5\baselineskip,
  belowskip = 0.5\baselineskip,]
CONSTRAINTS = '\n'.join(
    f"{i+1}. Include the exact word: '{constraint}'."
    for i, constraint in enumerate(constraints)
)
\end{lstlisting}

\clearpage
\section{Keyword Instructions}\label{app:vocab_table}
\begin{table}[htbp]
\centering
\tiny
\caption{Complete vocabulary of 500 business-relevant terms extracted from SEC 10-K filings, ranked by generation difficulty and used as instruction constraints in IFScale. Terms span from simple business concepts to complex technical terminology, ensuring varying difficulty in adherence across instructions.}
\vspace{0.5em}
\begin{tabular}{llllllll}
\toprule
ESG & ROI & debt & edge & tort & vest & chain & china \\
churn & cloud & cycle & fixed & fleet & goods & gross & HIPAA \\
japan & joint & labor & legal & patch & pivot & proxy & range \\
rural & sheet & shelf & solar & spark & stack & stock & union \\
yield & EBITDA & active & annual & appeal & backed & collar & common \\
cortex & credit & crypto & decree & design & diesel & direct & energy \\
equity & ethics & europe & export & factor & filing & fiscal & frills \\
frozen & future & global & hazard & health & hybrid & import & inputs \\
issuer & lessor & linear & merger & mobile & modify & neural & online \\
parent & patent & payout & rebate & recall & return & safety & salary \\
select & states & survey & talent & tariff & ticket & treaty & trends \\
united & volume & voting & wealth & trustee & adverse & battery & biotech \\
captive & charter & climate & conduct & consent & content & council & defense \\
digital & diluted & economy & entries & exhibit & expense & exploit & fintech \\
general & greater & holders & holding & insider & insight & interim & journal \\
journey & justice & latency & loyalty & meeting & modular & netting & offices \\
opinion & optical & organic & payroll & pioneer & premium & product & protein \\
quantum & quarter & reality & repairs & revenue & roadmap & rollout & salvage \\
seating & secrets & startup & subsidy & summary & surplus & tiering & tuition \\
upgrade & venture & virtual & website & willful & working & adequacy & advisory \\
affinity & alliance & argument & blackout & breaches & briefing & bundling & callable \\
cashflow & chargers & clinical & conflict & covenant & currency & delaware & director \\
distinct & dividend & domestic & downtime & emerging & emphasis & endpoint & enrollee \\
episodic & estimate & evidence & exposure & facility & floating & forensic & hydrogen \\
indirect & industry & issuance & judgment & leverage & magazine & majority & mandates \\
matching & material & maturing & mismatch & mobility & monopoly & mortgage & overhead \\
pharmacy & pipeline & platform & pursuant & reckless & research & residual & response \\
retrofit & robotics & sabotage & sanction & scalable & seamless & shutdown & solvency \\
spectrum & sporting & staffing & standard & stations & subpoena & taxonomy & traction \\
turnover & upstream & wellness & actuarial & adaptable & agreement & algorithm & amendment \\
amenities & anonymity & antitrust & appraisal & attrition & augmented & autopilot & bandwidth \\
beverages & borrowing & budgeting & complaint & completed & container & copayment & copyright \\
detection & discovery & downgrade & ecommerce & ecosystem & emergency & endeavors & executive \\
expansion & expertise & extension & fiduciary & financial & formulary & franchise & frequency \\
grounding & headcount & hierarchy & incentive & inclusion & indenture & integrity & intensive \\
liquidity & logistics & mainframe & mechanism & mediation & messaging & milestone & municipal \\
narrative & nonpublic & nutrition & objective & occupancy & oversight & paragraph & penalties \\
portfolio & proposals & provision & publisher & qualified & reference & reimburse & retention \\
revolving & royalties & scorecard & severance & shortages & showrooms & signature & solutions \\
sovereign & specialty & stainless & strategic & streaming & synergies & telephone & trademark \\
treatment & uninsured & wholesale & artificial & assumption & bankruptcy & bottleneck & capitation \\
collateral & colocation & commercial & competitor & compromise & confirming & creativity & deductible \\
defamation & definitive & department & derivative & discussion & durability & encryption & engagement \\
escalation & experience & forfeiture & generative & geothermal & governance & healthcare & households \\
impairment & impression & initiative & injunction & innovation & leadership & marketable & maturities \\
measurable & mitigation & moderation & multimodal & negligence & nomination & noncurrent & observable \\
onboarding & permitting & philosophy & powertrain & prevention & principles & proceeding & processing \\
projection & promissory & properties & prospectus & protection & redemption & redundancy & remittance \\
resilience & resolution & securities & settlement & strategies & subsidiary & succession & technology \\
theatrical & throughput & timeliness & vertically & washington & arbitration & arrangement & attractions \\
attribution & centralized & commodities & comparative & competition & composition & computation & consumables \\
convenience & convergence & correlation & deliverable & demographic & divestiture & eligibility & enforcement \\
enhancement & equivalents & expenditure & facultative & foreclosure & fulfillment & geographies & gigafactory \\
information & inventories & legislation & liabilities & liquidation & maintenance & materiality & opportunity \\
origination & outstanding & partnership & penetration & perishables & recognition & recruitment & remediation \\
seasonality & sensitivity & simulations & statistical & stewardship & subrogation & supercenter & translation \\
unqualified & utilization & withholding & agricultural & architecture & compensation & contribution & dispositions \\
distillation & facilitation & installation & intellectual & intelligence & intercompany & localization & monetization \\
multilateral & neighborhood & obsolescence & optimization & policyholder & presentation & productivity & proportional \\
repatriation & supercharger & transmission & transparency & undiscounted & unobservable & affordability & collaboration \\
comprehensive & concentration & disagreements & entertainment & hydroelectric & international & macroeconomic & remeasurement \\
vulnerability & whistleblower & administrative & infrastructure & noncontrolling & reconciliation & sustainability & transportation \\
unconventional & personalization & electromyography & commercialization & & & & \\
\bottomrule
\end{tabular}
\label{tab:vocabulary}
\end{table}

\end{document}